\documentclass{article}

\PassOptionsToPackage{numbers,sort&compress}{natbib}

\usepackage[main, final]{neurips_2026}

\usepackage[utf8]{inputenc} %
\usepackage[T1]{fontenc}    %
\usepackage{url}            %
\usepackage{graphicx}
\usepackage{booktabs}       %
\usepackage{amsfonts}       %
\usepackage{amssymb}        %
\usepackage{amsmath}
\usepackage{nicefrac}       %
\usepackage{microtype}      %
\usepackage[table]{xcolor} %
\usepackage{multirow}
\usepackage{pifont}
\usepackage{caption}
\usepackage{float}
\usepackage{wrapfig}
\usepackage{placeins}
\usepackage{CJKutf8}
\usepackage[most]{tcolorbox}
\usepackage{algorithm}
\usepackage[noend]{algpseudocode}

\usepackage{hyperref}       %
\hypersetup{
    colorlinks=true,
    linkcolor=[rgb]{0,0,0.6},
    citecolor=[rgb]{0,0,0.6},
    filecolor=[rgb]{0,0,0.6},
    urlcolor=[rgb]{0,0,0.6}
}

\title{Vision-OPD: Learning to See Fine Details for Multimodal LLMs via On-Policy Self-Distillation}

\author{
  Qianhao Yuan$^{1,2}$, Jie Lou$^{3,\dagger}$, Xing Yu$^{3}$, Hongyu Lin$^{1}$, Le Sun$^{1}$, \textbf{Xianpei Han$^{1}$,} \textbf{Yaojie Lu$^{1,\dagger}$}  \\[2pt]
  $^{1}$Chinese Information Processing Laboratory, Institute of Software, Chinese Academy of Sciences \\[2pt]
  $^{2}$University of Chinese Academy of Sciences $^{3}$Xiaohongshu Inc. \\[2pt]
  \texttt{yuanqianhao2024@iscas.ac.cn }  \texttt{loujie0822@gmail.com } 
  \texttt{luyaojie@iscas.ac.cn} 
}

\begin{document}

\maketitle
\renewcommand{\thefootnote}{\arabic{footnote}}
\renewcommand{\thefootnote}{$\dagger$}
\footnotetext{Corresponding authors.}

\begin{abstract}
   Multimodal Large Language Models (MLLMs) still struggle with fine-grained visual understanding, where answers often depend on small but decisive evidence in the full image. We observe a regional-to-global perception gap: the same MLLM answers fine-grained questions more accurately when conditioned on evidence-centered crops than on the corresponding full images, suggesting that many failures stem from difficulty to focus on relevant evidence rather than insufficient local recognition ability. Motivated by this observation, we propose Vision-OPD (Vision On-Policy Distillation), a regional-to-global self-distillation framework that transfers the model's own privileged regional perception to its full-image policy. 
   Vision-OPD instantiates two conditional policies from the same MLLM: a crop-conditioned teacher and a full-image-conditioned student. The student generates on-policy rollouts, and Vision-OPD minimizes token-level divergence between the teacher and student next-token distributions along these rollouts. 
   This enables the model to internalize the benefit of visual zooming.
The Vision-OPD training objective uses no external teacher,
ground-truth labels, or reward verifier, and the trained model
requires no visual tool use at inference.
   Experiments on multiple fine-grained visual understanding benchmarks show that Vision-OPD models achieve competitive or superior performance against much larger open-source, closed-source, and ``Thinking-with-Images'' agentic models.
   The code is available at \url{https://github.com/VisionOPD/Vision-OPD}.
\end{abstract}

\section{Introduction}

Multimodal Large Language Models (MLLMs) have demonstrated impressive capabilities in general visual understanding and reasoning~\cite{comanici2025gemini,singh2025openai,google2025gemini3,openai2025gpt5}. However, they still struggle with fine-grained visual understanding, where the answer often depends on small but decisive details that occupy only a fraction of the image~\cite{zhang2025mllms,wang2025grasp,liu2025vlm}. In full-image inference, these details are often easy to overlook amid many visual tokens. As a result, MLLMs may produce plausible answers based on the global scene while missing the local evidence that is truly needed for the question.

Recent ``Thinking-with-Images'' methods~\cite{zheng2025deepeyes,zhang2025thyme} address this issue by equipping MLLMs with agentic visual tool use, enabling them to crop, zoom, and inspect the region of interest during inference. By making local evidence more salient, these methods improve fine-grained visual understanding. However, they introduce substantial inference overhead due to repeated image encoding and model calls. A natural question arises: \textit{can the benefit of visual zooming be internalized through training, so that the model can use fine-grained evidence from the full image without additional tool use?}

Our motivation comes from a simple observation: the same MLLM often answers a fine-grained question more accurately when conditioned on the evidence-centered crop than on the corresponding full image. 
This performance gap reveals a broader bottleneck: MLLMs can often interpret the relevant evidence once it is made salient, but struggle to exploit it when it is embedded in the global visual context. This observation suggests a natural training signal: the model's own crop-conditioned behavior can serve as privileged supervision for improving its full-image behavior.

A straightforward way to exploit such privileged supervision is supervised fine-tuning (SFT) on crop-conditioned responses. However, this trains the model on trajectories generated under privileged crop inputs, leading to distribution mismatch and exposure bias~\cite{agarwal2024policy}. Reinforcement learning with verifiable rewards methods, such as GRPO~\cite{shao2024deepseekmath} and DAPO~\cite{yu2025dapo}, can optimize on-policy rollouts, but usually provide only sparse sequence-level feedback, and require ground-truth labels and verifiers. On-Policy Distillation (OPD)~\cite{agarwal2024policy,lu2025onpolicydistillation} combines on-policy sampling with dense token-level supervision, but existing OPD methods typically rely on external stronger teachers or ground-truth label~\cite{zhao2026self}.

We propose Vision-OPD, a regional-to-global self-distillation framework for fine-grained visual understanding. It instantiates two policies from the same MLLM with different visual conditions: a crop-conditioned teacher that observes the evidence-centered crop as a privileged input, and a full-image-conditioned student that observes the full image as in standard inference. The student first generates on-policy rollouts from the full image. For each student-generated prefix, Vision-OPD computes the logit distributions of both the crop-conditioned teacher and the full-image-conditioned student, and minimizes their divergence. 
In this way, the model transfers its own privileged crop-conditioned
behavior to its full-image-conditioned policy along the student's
generation trajectory. The Vision-OPD training objective uses no
external teacher, ground-truth labels, or reward verifier, and the
trained model requires no visual tool use at inference.

We conduct extensive experiments to validate the effectiveness of Vision-OPD.
With only 6.2K synthetic training data, Vision-OPD enables a 9B model to outperform much larger open-source models (e.g., Qwen3.5-397B), closed-source models (e.g., GPT-5.4, Gemini-3.1-Pro, Gemini-3.5-Flash), and agentic ``Thinking-with-Images'' methods on fine-grained visual understanding tasks that require dense local evidence.
On hold-out tasks beyond the training distribution, Vision-OPD largely maintains general visual understanding and reasoning ability.
Further analyses verify the necessity of on-policy sampling and dense token-level supervision, and show that Vision-OPD substantially narrows the regional-to-global perception gap in MLLMs.

Our contributions are summarized as follows:
\begin{itemize}
\item We introduce a regional-to-global self-distillation formulation for fine-grained visual understanding, with privileged crop-conditioned behavior as supervision for full-image inference.
\item We propose Vision-OPD, an on-policy self-distillation framework where a crop-conditioned policy supervises a full-image policy on the student's rollouts via token-level supervision.
\item Comprehensive experiments validate the effectiveness of Vision-OPD. We demonstrate that Vision-OPD can significantly narrow the regional-to-global gap, and the on-policy sampling and dense supervision are important to its success.
\end{itemize}

\section{Preliminary: distillation and on-policy distillation (OPD)}

Knowledge distillation transfers the behavior of a strong teacher into a weaker student by matching the teacher's distributions.
For an input \(x\) and output sequence \(y=(y_1,\ldots,y_{|y|})\), traditional supervised distillation is off-policy: the student is trained on teacher-induced prefixes \(y_{<t}\) and minimizes
\begin{equation}
\mathcal{L}_{\mathrm{Supervised\ Distillation}}(\theta)=\mathbb{E}_{(x,y)\sim\mathcal{S}}\left[\frac{1}{|y|}\sum_{t=1}^{|y|} D\left(p_T(\cdot \mid y_{<t},x)\,\|\,p_S(\cdot \mid y_{<t},x)\right)\right].
\end{equation}
This supervised distillation objective provides dense token-level supervision, but at inference time the model conditions on its own prefixes rather than those observed in \(\mathcal{S}\), which creates a state-distribution mismatch and can cause errors to compound over long horizons~\citep{ross2011reduction}.

On-policy distillation (OPD) mitigates this mismatch by sampling \(y\sim p_S(\cdot\mid x)\) from the current student and querying the teacher on the prefixes generated by the student~\citep{agarwal2024policy,lu2025onpolicydistillation,xu2024speculative}. The objective is
\begin{equation}
\mathcal{L}_{\mathrm{OPD}}(\theta)=\mathbb{E}_{x\sim\mathcal{S},\,\hat{y}\sim p_S(\cdot\mid x)}\left[\frac{1}{|\hat{y}|}\sum_{t=1}^{|\hat{y}|} D\left(p_T(\cdot \mid \hat{y}_{<t},x)\,\|\,p_S(\cdot \mid \hat{y}_{<t},x)\right)\right].
\end{equation}
OPD combines the on-policy relevance of reinforcement learning with the dense token-level guidance of distillation, making it a natural fit for autoregressive generation and reasoning.

\begin{figure*}[t]
  \centering
  \begin{minipage}[t]{0.49\textwidth}
    \vspace{0pt}
    \centering
    \includegraphics[width=\linewidth]{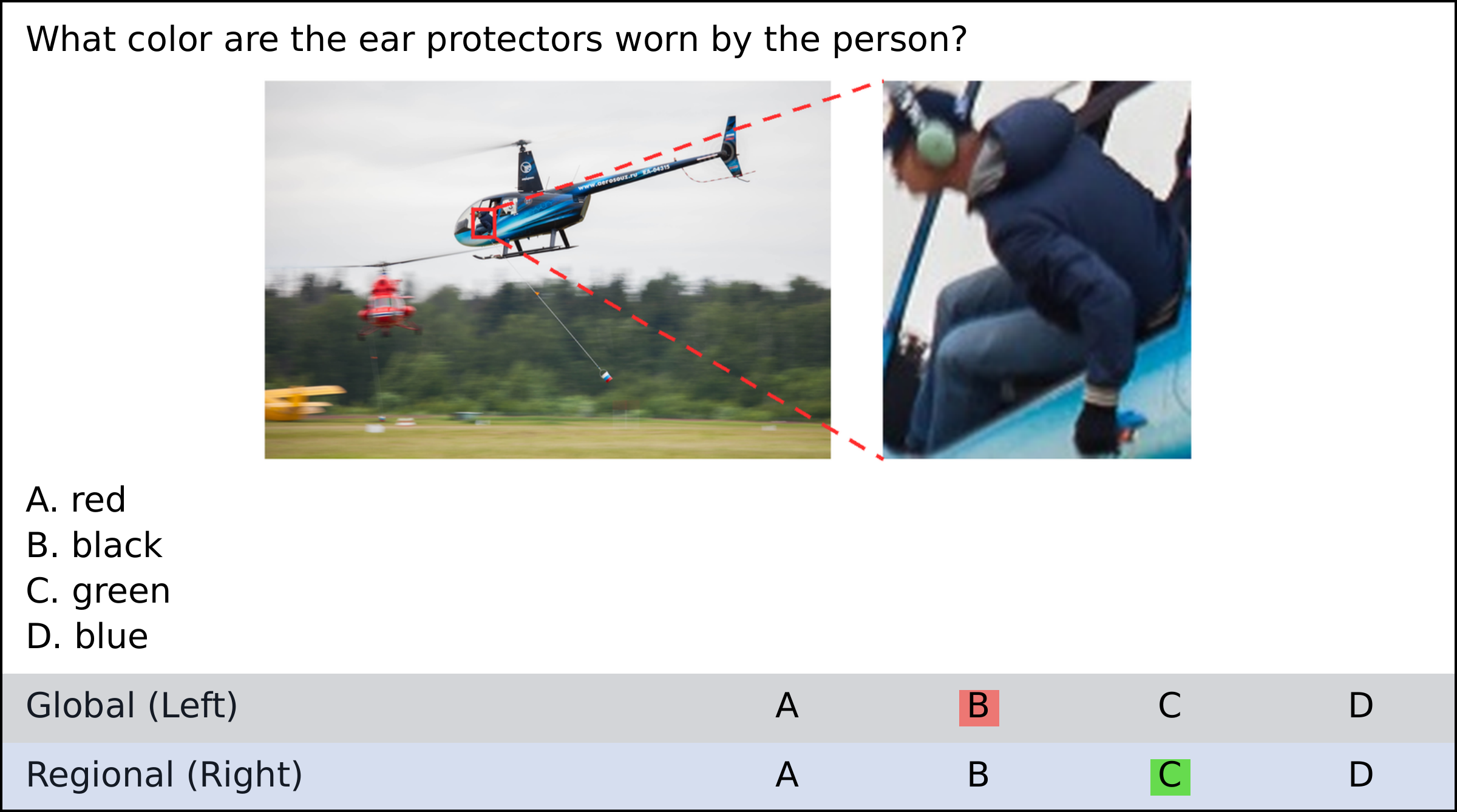}
    \captionof{figure}{A case of the regional-to-global gap, based on Qwen3.5-9B. The global image input leads to the wrong answer, while the cropped region input yields the correct answer.}
    \label{fig:regional_global_gap_left}
  \end{minipage}\hfill
  \begin{minipage}[t]{0.49\textwidth}
    \vspace{0pt}
    \centering
    \includegraphics[width=\linewidth]{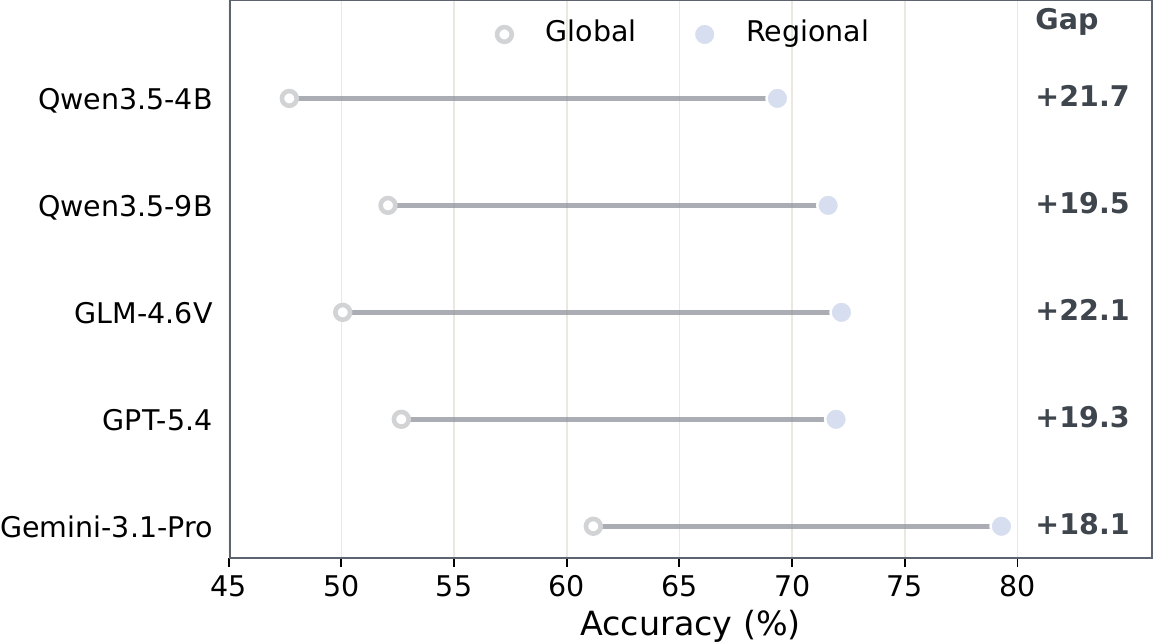}
    \captionof{figure}{The accuracy of regional inputs is consistently higher than the accuracy of global inputs, revealing a persistent regional-to-global gap across multiple MLLMs.}
    \label{fig:regional_global_gap_right}
  \end{minipage}
\end{figure*}

\FloatBarrier
\section{Vision-OPD}
\label{sec:method}
\subsection{Motivation: less is more for fine-grained visual understanding}
\label{sec:motivation}

MLLMs struggle with fine-grained visual understanding where small but decisive evidence is overwhelmed by global context. Recent ``Thinking-with-Images'' methods~\cite{zheng2025deepeyes,zhang2025thyme,chng2025sensenova} demonstrate that zooming into the region of interest improves fine-grained understanding, indicating that the bottleneck lies not in the model's recognition ability, but in its capacity to focus on relevant evidence within the full image.
This can be verified by comparing the model's performance under two conditions: when it receives the full image (global input) versus when it receives only an evidence-centered crop (regional input). If the model answers correctly with the crop but fails with the full image, it indicates that the model can recognize the evidence but struggles to focus on it within the global context. 

Figure~\ref{fig:regional_global_gap_left} illustrates a qualitative case on Qwen3.5-9B. The question asks about the ear-protector color. Given the full image, the model predicts black; given only the cropped region, it correctly predicts green. The decisive evidence is recognizable in isolation, but is overwhelmed by the global context.
Evaluation on ZoomBench~\citep{wei2026zooming} confirms that this pattern is systematic.
As shown in Figure~\ref{fig:regional_global_gap_right}, regional-input accuracy consistently exceeds global-input accuracy by 18--22 points. 
Even much larger models and closed-source models, such as GLM-4.6V, GPT-5.4, and Gemini-3.1-Pro, exhibit substantial gaps, 
confirming that parameter scaling alone does not close this regional-to-global gap. 

This observation motivates Vision-OPD: since a model's own regional perception consistently outperforms its global perception, we can use the former as privileged supervision for the latter, internalizing the benefits of zooming into a single forward pass, without inference-time tool use.

\subsection{Method: regional-to-global on-policy self-distillation}
\label{sec:opd_method}

\begin{figure*}[t]
  \centering
  \includegraphics[width=\textwidth]{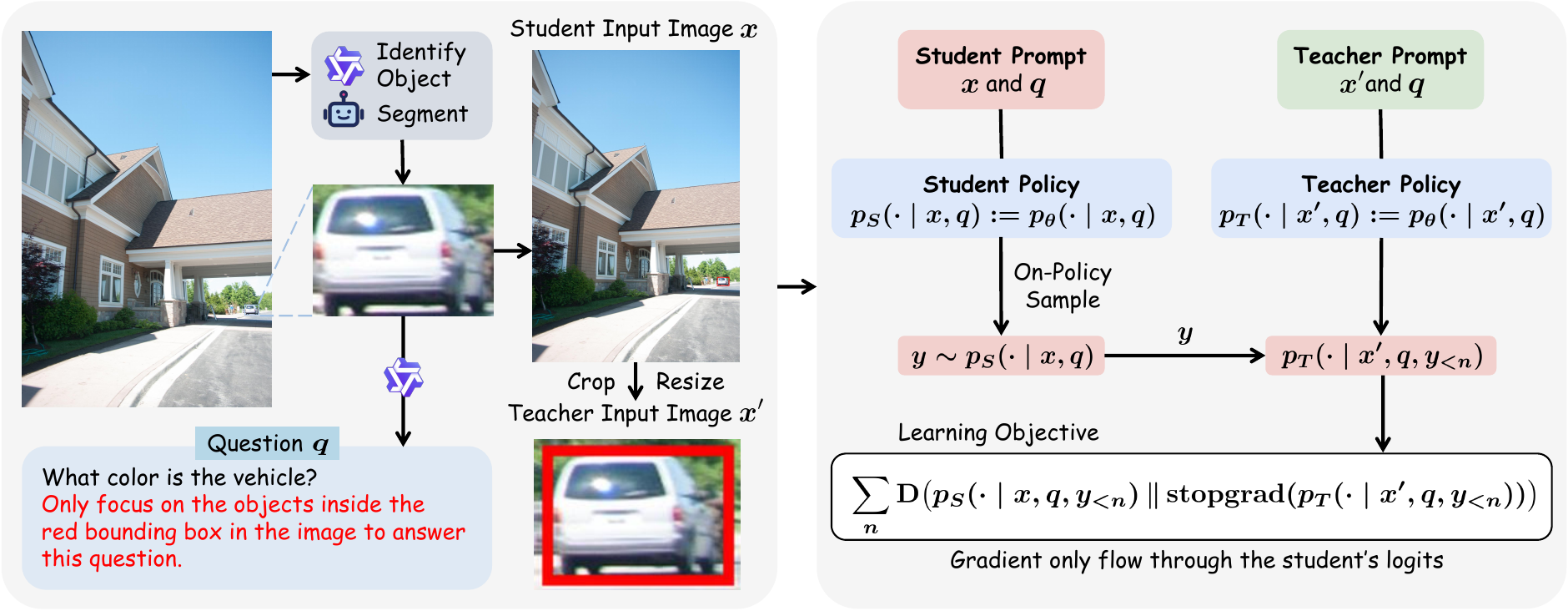}
  \caption{Overview of Vision-OPD. \textbf{Left:} Fine-grained visual questions are generated on evidence-centered crops and grounded back to the full image via bounding-box overlay. \textbf{Right:} A teacher policy \(p_T(\cdot \mid x_{\text{crop}})\) and a student policy \(p_S(\cdot \mid x_{\text{global}})\) are instantiated from the same MLLM. The student generates on-policy rollouts \(y \sim p_S\), and the per-token divergence \(D(p_T \| p_S)\) along these rollouts provides dense supervision. 
  Gradients flow only through the student's logits, enabling
self-distillation without ground-truth labels in the training
objective.
  }
  \label{fig:vision_opd_overview}
\end{figure*}

To exploit the regional-to-global gap for training, we construct a dataset of triplets \(\mathcal{D} = \{(x_i, x'_i, q_i)\}_{i=1}^{N}\), where each triplet pairs a full image \(x\) with its evidence-centered crop \(x'\) and a fine-grained question \(q\). As illustrated in Figure~\ref{fig:vision_opd_overview} (left), inspired by recent work on multimodal data synthesis~\citep{wei2026zooming,wang2026hopchain}, we first apply object identification and segmentation to propose bounding box based on a raw image \(I\), retaining only small regions (area ratio \(< \tau\)) that are likely to contain fine-grained evidence hidden in the image. For each retained region \(R\), we use Qwen3.5-397B~\citep{qwen3.5} as a question generator to produce a question \(q\) that are answerable from \(R\) alone.
To ground the question back to the full image and avoid referential ambiguity, the bounding box of \(R\) is overlaid onto \(I\) to produce \(x\), and a spatial constraint is appended to \(q\) (e.g., ``\textit{Only focus on the objects inside the red bounding box}''). 
Then we crop \(I\) to the bounding box of \(R\) and resize it by \(2\times\) to produce \(x'\).

Each resulting triplet \((x, x', q)\) thus presents the same question under two visual conditions: the student sees the full image with spatial guidance, while the teacher sees only the isolated crop. The gap between the two conditions directly serves as the learning signal for self-distillation. 
To make comparison with alternative training strategies, such as off-policy distillation (SFT), RLVR and OPSD~\cite{zhao2026self}, we also utilize Qwen3.5-397B as an answer generator to generate ground-truth labels for these methods. We sample multiple responses by giving the region \(R\) as image input, and keep a question only when the majority answer reaches a strict consensus (\(>0.75\)).
In total, we synthesize 6.2K samples for training.

Given the synthesized dataset, we now describe how Vision-OPD leverages these paired views for training (Figure~\ref{fig:vision_opd_overview}, right), with a complete summary in Algorithm~\ref{alg:vision_opd}.
We instantiate two conditional distributions from the same MLLM \(p_\theta\) by varying the visual conditioning context. The teacher policy conditions on the evidence-centered crop \(x'\) as privileged visual information, \(p_T(\cdot \mid x', q) = p_\theta(\cdot \mid x', q)\). The student policy observes the full image \(x\) and the question \(q\), \(p_S(\cdot \mid x, q) = p_\theta(\cdot \mid x, q)\). 
The student sees the global scene, while the teacher sees only the zoomed-in crop where the evidence is unambiguous, which grants the teacher a privileged view in which fine-grained details are isolated. 

Given a training sample \((x, x', q)\), the student generates an on-policy response \(y = (y_1, \ldots, y_{|y|}) \sim p_S(\cdot \mid x, q)\). Both policies then evaluate this student-generated trajectory. At each position \(n\), they induce next-token distributions over \(y_n \in \mathcal{V}\) conditioned on the same student prefix \(y_{<n} = (y_1, \ldots, y_{n-1})\): \(p_S(y_n \mid x, q, y_{<n})\) and \(p_T(y_n \mid x', q, y_{<n})\). By re-evaluating the same trajectory under a cleaner local view, the teacher's token-level distribution naturally encodes sharper attention to fine-grained visual evidence without any additional decoding.

The training objective minimizes the expected per-token divergence between teacher and student over the student's own rollouts. Given a student-generated sequence \(y\), the averaged divergence is
\begin{equation}
D(p_T \| p_S)(y \mid x, x', q) \;=\; \frac{1}{|y|}\sum_{n=1}^{|y|} D\!\Big(p_T(\cdot \mid x', q, y_{<n}) \,\Big\|\, p_S(\cdot \mid x, q, y_{<n})\Big),
\end{equation}
where \(D\) can be any distribution divergence such as the generalized Jensen--Shannon divergence \(\mathrm{JSD}_\beta\) for a weight \(\beta \in [0,1]\). It is defined as \(\mathrm{JSD}_\beta(p_T \| p_S) = \beta\, D_{KL}(p_T \| m) + (1-\beta)\, D_{KL}(p_S \| m)\), where \(m = \beta\, p_T + (1-\beta)\, p_S\) is the interpolated mixture. The overall loss minimizes the expected divergence over on-policy student samples:
\begin{equation}
\mathcal{L}_{\mathrm{Vision\text{-}OPD}}(\theta) = \mathbb{E}_{(x,x',q)\sim\mathcal{D}}\!\left[\mathbb{E}_{y \sim p_S(\cdot \mid x, q)}\!\left[D(p_T \| p_S)(y \mid x, x', q)\right]\right].
\end{equation}
Gradients are backpropagated only through the student policy \(p_S\), while the teacher \(p_T\) acts as a fixed target conditioned on the privileged crop \((x', q)\). Using the student's own generated sequence \(y\) as the training prefix aligns the state distribution between training and inference. This avoids the compounding errors that arise from prefix mismatch in off-policy distillation~\citep{ross2011reduction}. The dense per-token divergence signal contrasts with the sparse binary rewards of RLVR methods, such as GRPO and DAPO. In Vision-OPD, every token receives a meaningful gradient, so training does not stall when all samples in a batch happen to be correct or incorrect.

Taken together, Vision-OPD combines on-policy sampling with dense
token-level supervision. Its training objective uses no external
teacher, ground-truth labels, or verifier.
Because the crop \(x'\) is extracted from unlabeled images by a fully automated data synthesis pipeline, the method is compatible with arbitrary image corpora and internalizes fine-grained visual understanding into a single forward pass.

\begin{algorithm}[t]
\caption{Vision-OPD}
\label{alg:vision_opd}
\begin{algorithmic}[1]
\Require Training dataset \(\mathcal{D} = \{(x_i, x'_i, q_i)\}_{i=1}^{N}\); MLLM \(p_\theta\); divergence \(D\) (e.g., \(\mathrm{JSD}_\beta\))
\State Let \(p_S(\cdot \mid x, q) \!:=\! p_\theta(\cdot \mid x, q)\) and \(p_T(\cdot \mid x', q) \!:=\! p_\theta(\cdot \mid x', q)\) be the same model \(p_\theta\) under different visual conditioning.
\For{\(\mathrm{step} = 1, \ldots, M\)}
\State Sample a batch \(\mathcal{B} \subset \mathcal{D}\)
\ForAll{\((x, x', q) \in \mathcal{B}\)}
\State Sample on-policy response \(y \sim p_S(\cdot \mid x, q)\)
\State Compute the token-wise divergence along the student rollout:
\[
\ell(x, x', q) \leftarrow \frac{1}{|y|}\sum_{n=1}^{|y|} D\!\Big(p_S(\cdot \mid x, q, y_{<n}) \,\Big\|\, \mathrm{stopgrad}\big(p_T(\cdot \mid x', q, y_{<n})\big)\Big)
\]
\EndFor
\State Calculate loss \(\mathcal{L}_{\mathrm{Vision\text{-}OPD}}(\theta) \leftarrow \frac{1}{|\mathcal{B}|}\sum_{(x,x',q) \in \mathcal{B}} \ell(x, x', q)\) and update \(\theta\)
\EndFor
\end{algorithmic}
\end{algorithm}

\section{Experiments}

\subsection{Experimental settings}

\textbf{Model training.}
We apply Vision-OPD to Qwen3.5-4B/9B~\cite{qwen3.5}, with our 6.2K synthetic data.
We use JSD (\(\beta=0.5\)) as the divergence objective, and approximate the divergence via top-\(K\) distillation, which computes only the top-\(K\) logits of the student and the corresponding teacher logits, alongside a tail-probability term.
With \(K=100\), this avoids the high memory overhead of full-vocabulary logit distillation~\cite{zhao2026self} while capturing most of the distributional information, since the tokens beyond the
top-100 typically cover less than \(1\times10^{-13}\) of the probability mass in our scenario.
The teacher is regularized via exponential moving average (EMA).
We set the maximum on-policy generation length to 1024, and training epoch to 1.

\textbf{Benchmarks.}
We evaluate on two groups of benchmarks.
The first group targets fine-grained visual understanding: V*~Bench~\cite{wu2024v} measures the accuracy of locating and recognizing tiny visual targets within complex scenes; ZoomBench~\cite{wei2026zooming} requires models to answer questions that depend on details at varying zoom levels; HR~Bench~\cite{wang2025divide} evaluates high-resolution perception at two resolutions (4K and 8K); and MME-RealWorld~\cite{zhang2024mme} covers real-world scenarios with high-resolution photographs.
The second group serves as holdout tasks for evaluating generalization beyond the training distribution, including MMVP~\cite{tong2024eyes}, CV-Bench~\cite{tong2024cambrian}, MMStar~\cite{chen2024we}, and POPE~\cite{li2023evaluating}.
These holdout benchmarks measure whether models retain general multimodal capabilities after fine-grained specialization.

\textbf{Baselines.}
We compare against two groups of baselines.
The first group compares Vision-OPD with existing SOTA models: (a)~``Thinking-with-Images'' agentic models that crop and zoom into the image region through multi-step reasoning, including DeepEyes~\cite{zheng2025deepeyes}, Thyme~\cite{zhang2025thyme}, DeepEyesV2~\cite{hong2025deepeyesv2}, and SenseNova-MARS~\cite{chng2025sensenova}; (b)~closed-source models, including GPT-5.2~\cite{openai2025gpt52}, GPT-5.4~\cite{openai2026gpt54}, Gemini-3.5-Flash~\cite{google2026gemini35}, and Gemini-3.1-Pro~\cite{google2026gemini3}; and (c)~open-source models of varying scales, including MiMo-VL-7B-RL~\cite{coreteam2025mimovltechnicalreport}, Qwen3-VL-Instruct~\cite{bai2025qwen3}, ZwZ~\cite{wei2026zooming}, MiniCPM-V-4.5~\cite{yu2025minicpm}, GLM-4.6V~\cite{hong2025glm}, Qwen3.5~\cite{qwen3.5}, and Kimi-K2.6~\cite{team2026kimi26}.
For GPT models, input images exceeding the API's resolution limit
were downscaled only as much as necessary to satisfy the limit, while
all other evaluation settings were kept identical across models.
The second group evaluates alternative training strategies under the same data and backbones: (a)~SFT on self-teacher, which performs supervised fine-tuning (SFT) on successful generations from the self-teacher, acting as an off-policy distillation baseline; (b)~Reinforcement Learning (RL) methods, including GRPO~\cite{shao2024deepseekmath} and DAPO~\cite{yu2025dapo}, which optimize via group relative policy gradient with binary outcome rewards verified against ground-truth answers; and (c)~OPSD~\cite{zhao2026self}, which uses on-policy self-distillation with ground-truth labels to provide reward signals.
All training-strategy baselines use the same data as Vision-OPD.
For SFT, RL and OPSD, the ground-truth labels used are generated as described in Section~\ref{sec:opd_method}.
Besides, we all use the non-thinking mode of Qwen3.5 models with different sizes (4B, 9B, 397B) for training and evaluation.

\begin{table*}[t]
\caption{Comparison with SOTA MLLMs.
We report accuracy (\%) for each model. Among open-source models (single forward pass), the best results are highlighted in \textbf{bold}, and the second-best are \underline{underlined}. Vision-OPD achieving the best overall performance.
}
\label{tab:experiments_main_results}
\centering
\scriptsize
\setlength{\tabcolsep}{3pt}
\resizebox{\textwidth}{!}{%
\begin{tabular}{lcccccccc}
\toprule
\multirow{2}{*}{\textbf{Model}} & \multirow{2}{*}{\shortstack{\textbf{Param}\\\textbf{Size}}} & \multirow{2}{*}{\textbf{V* Bench}} & \multirow{2}{*}{\textbf{ZoomBench}} & \textbf{HR-Bench} & \textbf{HR-Bench} & \textbf{MME-RW} & \textbf{MME-RW} & \multirow{2}{*}{\textbf{Average}} \\
 &  &  &  & \textbf{4K} & \textbf{8K} & \textbf{EN} & \textbf{CN} &  \\
\midrule
\multicolumn{9}{c}{\textbf{\textit{``Thinking-with-Images'' Agentic Models}}} \\
\midrule
DeepEyes & 7B & 85.86 & 46.51 & 75.13 & 72.63 & 64.10 & 64.09 & 68.05 \\
Thyme & 7B & 82.20 & 45.09 & 77.00 & 72.00 & 64.80 & 64.59 & 67.61 \\
DeepEyesV2 & 7B & 81.68 & 44.97 & 77.88 & 73.75 & 64.90 & 65.07 & 68.04 \\
SenseNova-MARS & 8B & 92.15 & 47.81 & 83.13 & 78.38 & 67.90 & 68.90 & 73.05 \\
\midrule
\multicolumn{9}{c}{\textbf{\textit{Closed-Source Models (Single Forward Pass)}}} \\
\midrule
GPT-5.2 & - & 79.06 & 50.89 & 80.88 & 78.25 & 72.37 & 68.51 & 71.66 \\
GPT-5.4 & - & 76.96 & 52.66 & 83.75 & 77.25 & 74.17 & 70.64 & 72.57 \\
Gemini-3.1-Pro & - & 86.91 & 61.18 & 89.63 & 86.50 & 74.22 & 72.05 & 78.41 \\
Gemini-3.5-Flash & - & 89.01 & 61.42 & 88.63 & 86.62 & 74.88 & 73.06 & 78.94 \\
\midrule
\multicolumn{9}{c}{\textbf{\textit{Open-Source Models (Single Forward Pass)}}} \\
\midrule
Qwen3.5 & 4B & 84.29 & 47.69 & 82.13 & 78.38 & 62.75 & 62.67 & 69.65 \\
Qwen3.5 & 9B & 82.72 & 52.07 & 84.38 & 79.13 & 70.31 & 67.30 & 72.65 \\
MiMo-VL-RL & 7B & 82.20 & 45.68 & 72.25 & 67.13 & 60.54 & 53.86 & 63.61 \\
Qwen3-VL-Instruct & 8B & 84.82 & 42.96 & 77.50 & 73.00 & 63.19 & 64.61 & 67.68 \\
ZwZ & 8B & 87.96 & 56.69 & 82.88 & 80.88 & 66.57 & 68.09 & 73.84 \\
MiniCPM-V-4.5 & 9B & 70.68 & 42.60 & 69.63 & 61.50 & 59.66 & 57.92 & 60.33 \\
GLM-4.6V & 106B & 86.91 & 50.06 & 79.63 & 76.13 & 65.57 & 63.36 & 70.28 \\
Qwen3-VL-Instruct & 235B & \underline{91.10} & 56.09 & 84.25 & 78.38 & 71.74 & 69.02 & 75.10 \\
Qwen3.5 & 397B & 87.96 & 57.16 & \textbf{88.25} & \underline{84.50} & \textbf{74.58} & \underline{69.26} & \underline{76.95} \\
Kimi-K2.6 & 1T & 88.48 & 53.14 & 81.63 & 78.00 & 68.84 & 66.11 & 72.70 \\
\midrule
\rowcolor{cyan!20}
Vision-OPD (Ours) & 4B & \underline{91.10} & \underline{59.76} & 82.25 & 78.63 & 71.13 & 68.78 & 75.28 \\
\rowcolor{cyan!20}
Vision-OPD (Ours) & 9B & \textbf{94.76} & \textbf{65.80} & \underline{86.88} & \textbf{84.63} & \underline{72.06} & \textbf{70.41} & \textbf{79.09} \\
\bottomrule
\end{tabular}%
}
\end{table*}

\subsection{Experimental results}
\subsubsection{Comparison with SOTA MLLMs}

\textbf{Performance gains over initial baselines.}
As shown in Table~\ref{tab:experiments_main_results}, Vision-OPD models consistently improve over the corresponding Qwen3.5 baselines across all benchmarks, demonstrating that Vision-OPD can effectively internalize fine-grained visual understanding capabilities into existing MLLMs.
See Appendix~\ref{app:case_study} for qualitative case study.

\textbf{Surpassing much larger open-source and closed-source models.}
Vision-OPD-9B surpasses open-source baselines regardless of scale, e.g., GLM-4.6V and Kimi-K2.6.
Compared with closed-source models, Vision-OPD-9B outperforms GPT-5.4 and even Gemini-3.1-Pro.

\textbf{Comparison with ``Thinking-with-Images'' agentic models.}
We further compare Vision-OPD with representative agentic models that explicitly zoom into image region during inference.
Despite requiring only a single forward pass, Vision-OPD models outperform these agentic models.

\begin{table*}[t]
\caption{Comparison with SFT, RLVR methods and OPSD. 
Vision-OPD consistently outperforms all baselines on fine-grained visual tasks while maintaining strong performance on holdout tasks.}
\label{tab:ood_generalization}
\centering
\resizebox{\textwidth}{!}{%
\begin{tabular}{lcccccccc}
\toprule
\multirow{3}{*}{\textbf{Method}} & \multicolumn{4}{c}{\textbf{Fine-Grained Visual Tasks}} & \multicolumn{4}{c}{\textbf{Holdout Tasks}} \\
\cmidrule(lr){2-5} \cmidrule(lr){6-9}
 & \multirow{2}{*}{\textbf{V* Bench}} & \multirow{2}{*}{\textbf{ZoomBench}} & \textbf{HR-Bench} & \textbf{HR-Bench} & \multirow{2}{*}{\textbf{MMVP}} & \multirow{2}{*}{\textbf{CV-Bench}} & \multirow{2}{*}{\textbf{MMStar}} & \multirow{2}{*}{\textbf{POPE}} \\
 &  &  & \textbf{4K} & \textbf{8K} &  &  &  &  \\
\midrule
\rowcolor{gray!20}
\rowcolor{gray!15}\multicolumn{9}{c}{\textit{Qwen3.5-4B}} \\
Vanilla & 84.29 & 47.69 & 82.13 & 78.38 & 76.67 & 86.51 & \textbf{72.07} & 88.28 \\
\midrule
SFT on Self-Teacher & 78.01 & 54.67 & 78.38 & 75.25 & 78.33 & 85.35 & 66.07 & 87.48 \\
GRPO & 83.77 & 55.38 & 81.50 & 77.38 & 79.33 & 87.20 & 70.07 & 86.37 \\
DAPO & 84.82 & 55.74 & 81.63 & 76.38 & 79.33 & 84.79 & 68.07 & 86.62 \\
OPSD & 81.68 & 53.85 & 80.38 & 76.63 & 78.68 & \textbf{87.27} & 67.53 & 88.83 \\
\rowcolor{cyan!20}
Vision-OPD (Ours) & \textbf{91.10} & \textbf{59.76} & \textbf{82.25} & \textbf{78.63} & \textbf{79.67} & \textbf{87.27} & 71.93 & \textbf{89.14} \\
\midrule
\rowcolor{gray!20}
\rowcolor{gray!15}\multicolumn{9}{c}{\textit{Qwen3.5-9B}} \\
Vanilla & 82.72 & 52.07 & 84.38 & 79.13 & 83.33 & 88.04 & \textbf{77.13} & 88.88 \\
\midrule
SFT on Self-Teacher & 82.20 & 58.46 & 81.88 & 78.88 & 80.67 & 87.45 & 69.07 & 87.47 \\
GRPO & 85.34 & 57.51 & 86.50 & 83.13 & 81.67 & 87.77 & 72.73 & 87.78 \\
DAPO & 88.48 & 55.62 & 86.25 & 83.88 & 79.33 & 87.30 & 75.67 & 87.54 \\
OPSD & 87.43 & 57.51 & 82.25 & 79.75 & 80.67 & 87.15 & 72.60 & 87.48 \\
\rowcolor{cyan!20}
Vision-OPD (Ours) & \textbf{94.76} & \textbf{65.80} & \textbf{86.88} & \textbf{84.63} & \textbf{83.67} & \textbf{88.07} & 76.53 & \textbf{89.13} \\
\bottomrule
\end{tabular}%
}
\end{table*}

\begin{table*}[t]
\caption{Comparison of various teacher regularization strategies, based on Qwen3.5-9B. Both trust-region regularization and exponential moving average (EMA) regularization use update coefficient \(\alpha = 0.05\). \(^\dagger\)Training with the current policy as teacher leads to collapse.}
\label{tab:teacher_ablation}
\centering
\scriptsize
\setlength{\tabcolsep}{3pt}
\resizebox{\textwidth}{!}{%
\begin{tabular}{lccccccc}
\toprule
\multirow{2}{*}{\textbf{Teacher}} & \multirow{2}{*}{\textbf{V* Bench}} & \multirow{2}{*}{\textbf{ZoomBench}} & \textbf{HR-Bench} & \textbf{HR-Bench} & \textbf{MME-RW} & \textbf{MME-RW} & \multirow{2}{*}{\textbf{Average}} \\
 &  &  & \textbf{4K} & \textbf{8K} & \textbf{EN} & \textbf{CN} &  \\
\midrule
Current Policy\(^\dagger\) & 0.00 & 0.00 & 0.00 & 0.00 & 3.51 & 0.05 & 0.59 \\
Initial Policy & 93.72 & 63.91 & 86.63 & 85.38 & 72.59 & 70.51 & 78.79 \\
Trust-Region Regularization & 93.19 & 63.79 & 87.38 & 85.13 & 72.31 & 70.75 & 78.76 \\
EMA Regularization & 94.76 & 65.80 & 86.88 & 84.63 & 72.06 & 70.41 & 79.09 \\
\bottomrule
\end{tabular}%
}
\end{table*}

\subsubsection{Comparison with SFT, RLVR methods and OPSD}

As shown in Table~\ref{tab:ood_generalization}, Vision-OPD consistently outperforms all alternative training strategies on fine-grained visual understanding benchmarks.
Another advantage of Vision-OPD is its ability to improve fine-grained capabilities while largely preserving previously acquired ones.
We evaluate this by testing on holdout benchmarks (MMVP, CV-Bench, MMStar, POPE), the distribution of which is unseen during training.
SFT on Self-Teacher exhibits severe forgetting, and RLVR methods (GRPO and DAPO) also degrade holdout performance.
In contrast, Vision-OPD largely maintains the capabilities on holdout performance.
These results suggest that Vision-OPD mitigates the performance--forgetting tradeoff observed in some alternative training approaches.

We further compare with OPSD~\cite{zhao2026self}, which also employs on-policy self-distillation but relies on ground-truth labels to provide reward signals.
Vision-OPD instead leverages a self-generated teacher that provides
dense token-level supervision without using ground-truth labels in
the distillation objective.
On fine-grained tasks, Vision-OPD outperforms OPSD on both scales.
Moreover, Vision-OPD achieves stronger holdout performance, confirming that our method leads to both better task performance and more robust generalization.

\subsection{Ablation study \& analysis}

\subsubsection{Effect of teacher regularization}

Vision-OPD uses a self-teacher that is initialized from the same checkpoint with the student and is updated throughout training. Proper regularization of the teacher is critical to prevent the teacher and student from co-adapting, which would cause training to collapse. We compare four teacher strategies in Table~\ref{tab:teacher_ablation}: (1) the current policy without any regularization, (2) the initial policy (frozen at initialization), (3) trust-region regularization, and (4) exponential moving average (EMA) regularization.

The experimental results are shown in Table~\ref{tab:teacher_ablation}.
Without regularization, training with the current policy as teacher diverges entirely, collapsing to near-zero accuracy across all benchmarks. This confirms that naive self-distillation without teacher regularization is fundamentally unstable. The initial policy, frozen at its pre-trained weights, already provides a strong teaching signal.
EMA regularization achieves the highest average.
Therefore, we adopt EMA regularization with update coefficient \(\alpha = 0.05\) for all remaining experiments.

\begin{table*}[t]
\caption{Comparison of divergence objectives, based on Qwen3.5-9B. We compare forward KL, reverse KL, and JSD (\(\beta=0.5\)) as the token-wise divergence measure \(D\).}
\label{tab:divergence_ablation}
\centering
\scriptsize
\setlength{\tabcolsep}{3pt}
\resizebox{\textwidth}{!}{%
\begin{tabular}{lccccccc}
\toprule
\multirow{2}{*}{\textbf{Method}} & \multirow{2}{*}{\textbf{V* Bench}} & \multirow{2}{*}{\textbf{ZoomBench}} & \textbf{HR-Bench} & \textbf{HR-Bench} & \textbf{MME-RW} & \textbf{MME-RW} & \multirow{2}{*}{\textbf{Average}} \\
 &  &  & \textbf{4K} & \textbf{8K} & \textbf{EN} & \textbf{CN} &  \\
\midrule
Forward KL (\(\mathrm{KL}(p_T \| p_S)\)) & 90.05 & 64.02 & 86.50 & 83.50 & 71.75 & 70.14 & 77.66 \\
Reverse KL (\(\mathrm{KL}(p_S \| p_T)\)) & 89.53 & 62.72 & 86.38 & 84.63 & 72.24 & 69.87 & 77.56 \\
JSD (\(\beta = 0.5\)) & 94.76 & 65.80 & 86.88 & 84.63 & 72.06 & 70.41 & 79.09 \\
\bottomrule
\end{tabular}%
}
\end{table*}

\begin{table*}[t]
\centering
\caption{Effect of generation length on Vision-OPD, based on Qwen3.5-9B. We compare on-policy sampled student generation lengths of 512 and 1024 tokens.}
\label{tab:generation_length}
\scriptsize
\setlength{\tabcolsep}{3pt}
\resizebox{\textwidth}{!}{%
\begin{tabular}{lccccccc}
\toprule
\multirow{2}{*}{\textbf{Generation Length}} & \multirow{2}{*}{\textbf{V* Bench}} & \multirow{2}{*}{\textbf{ZoomBench}} & \textbf{HR-Bench} & \textbf{HR-Bench} & \textbf{MME-RW} & \textbf{MME-RW} & \multirow{2}{*}{\textbf{Average}} \\
 &  &  & \textbf{4K} & \textbf{8K} & \textbf{EN} & \textbf{CN} &  \\
\midrule
512 Tokens & 91.62 & 64.38 & 87.25 & 85.00 & 71.62 & 69.44 & 78.22 \\
1024 Tokens & 94.76 & 65.80 & 86.88 & 84.63 & 72.06 & 70.41 & 79.09 \\
\bottomrule
\end{tabular}%
}

\vspace{2.4em}
\caption{Ablation on divergence computation strategies for Vision-OPD, based on Qwen3.5-9B.
Top-K logits distillation outperforms sampled-token objectives.
}
\label{tab:distillation_variant}
\scriptsize
\setlength{\tabcolsep}{3pt}
\resizebox{\textwidth}{!}{%
\begin{tabular}{lccccccc}
\toprule
\multirow{2}{*}{\textbf{Method}} & \multirow{2}{*}{\textbf{V* Bench}} & \multirow{2}{*}{\textbf{ZoomBench}} & \textbf{HR-Bench} & \textbf{HR-Bench} & \textbf{MME-RW} & \textbf{MME-RW} & \multirow{2}{*}{\textbf{Average}} \\
 &  &  & \textbf{4K} & \textbf{8K} & \textbf{EN} & \textbf{CN} &  \\
\midrule
Sampled-token distillation & 93.72 & 61.54 & 85.50 & 84.25 & 72.09 & 68.31 & 77.57 \\
Top-K logits distillation & 94.76 & 65.80 & 86.88 & 84.63 & 72.06 & 70.41 & 79.09 \\
\bottomrule
\end{tabular}%
}
\end{table*}

\subsubsection{Effect of divergence objective}

A design choice in Vision-OPD is the divergence used for per-token distribution matching between the teacher and the student. We compare forward KL, reverse KL, and JSD (\(\beta=0.5\)) with Qwen3.5-9B in Table~\ref{tab:divergence_ablation}. 
JSD (\(\beta=0.5\)) yields the strongest gains, outperforming forward KL and reverse KL. 
Therefore, we adopt JSD (\(\beta=0.5\)) in all remaining experiments.

\subsubsection{Effect of generation length}

Since our objective operates at the token level, the number of generated tokens per sample directly affects the amount of supervision signal available to the student. 
We compare different generation lengths on Qwen3.5-9B in Table~\ref{tab:generation_length}. Increasing the generation length from 512 to 1024 tokens yields performance improvements, suggesting that longer rollouts provide richer supervision for fine-grained visual understanding tasks. We adopt 1024 tokens for all remaining experiments.

\subsubsection{Learning objective comparison: top-K logits distillation vs. sampled-token distillation}

The learning objective of Vision-OPD is defined as a per-token discrepancy between the teacher and student distributions. 
We compare this objective in two ways: (1) Top-K logits distillation~\cite{hubotter2026reinforcement}: for each token position, we compute the divergence over the top-\(K\) logits via a partial softmax, yielding a proper token-level divergence between the two policies. Specifically, we retain only the top-\(K\) logits of the student logits and the corresponding teacher logits, complemented by a tail-probability term that accounts for the remaining probability mass. With \(K=100\), this could capture most of the distributional information, since the tokens beyond the
top-100 typically cover less than \(1\times10^{-13}\) of the probability mass in our scenario.
(2) Sampled-token policy-gradient objective~\cite{lu2025onpolicydistillation}: we evaluate teacher and student log-probabilities only at the token actually sampled by the student, and use the log-probability ratio between teacher and student as a scalar advantage inside a policy-gradient-style loss. 
This is analogous to how RLVR methods (e.g., GRPO) apply a constant scalar advantage to sampled tokens, but shaped by the teacher's log-probabilities rather than a binary reward signal.

We compare these variants on Qwen3.5-9B in Table~\ref{tab:distillation_variant}. Top-K logits distillation provides an overall performance gain over the sampled-token objective, confirming that dense logit-level credit assignment leads to more effective learning than scalar per-token shaping. Therefore, we adopt top-\(K\) logits distillation with \(K=100\) for all remaining experiments.

\subsubsection{Vision-OPD significantly narrows the regional-to-global gap}

As mentioned in Section~\ref{sec:motivation}, fine-grained failures often arise not because the decisive evidence 
\begin{wrapfigure}{r}{0.48\textwidth}
  \centering
  \includegraphics[width=\linewidth]{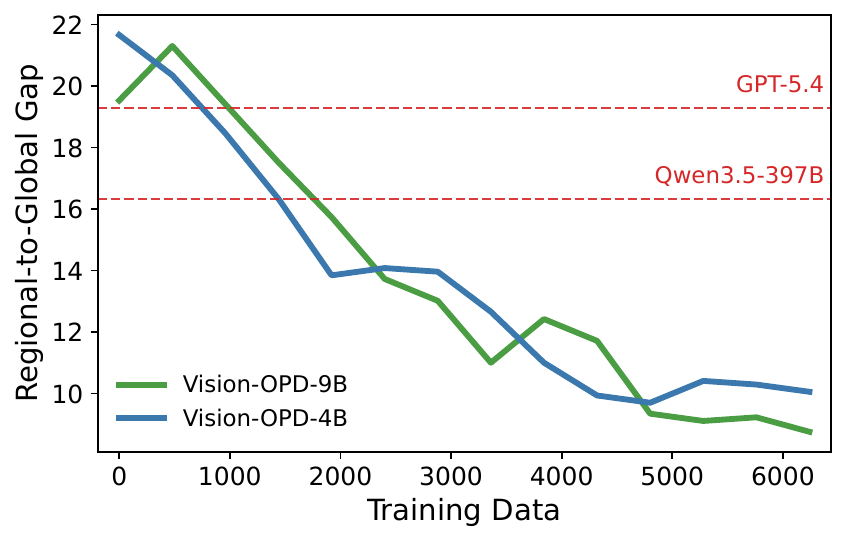}
  \captionof{figure}{Regional-to-global gap during Vision-OPD training. A lower gap indicates that the model can better recover crop-visible evidence from the full image.}
  \label{fig:regional_to_global_gap_training}
\end{wrapfigure}
is unrecognizable, but because it is hard for current MLLMs to focus on that evidence within the full image.
To test whether Vision-OPD addresses this bottleneck during training, we use the same comparison as in Section~\ref{sec:motivation}: each checkpoint answers the same question with the full image as input and with the evidence-centered crop as input.
We track the resulting regional-to-global gap over training, where a smaller gap indicates that evidence visible in the crop is being recovered more reliably from the full image.
Figure~\ref{fig:regional_to_global_gap_training} shows that Vision-OPD steadily closes this gap during training. 
This reduction is substantial: both Vision-OPD models end with a smaller gap than much larger or closed-source models.
These results demonstrate that through Vision-OPD, the models learn to focus on fine-grained evidence directly from the full image, internalizing the benefit of zooming without inference-time tool use.

\section{Related work}
\textbf{On-policy distillation.}
On-policy distillation~\cite{zhang2026shortopd,agarwal2024policy, lu2025onpolicydistillation, gu2024minillm,sun2026distill,yang2025qwen3} trains a student on its self-generated trajectories, with token-level teacher supervision via KL divergence or related objectives~\cite{sun2026simct,wang2026parabridge,ruan2026contrastive,wen2026constitutional,guan2026latent,li2026rethinking}.
This alignment reduces the train-test mismatch of off-policy distillation~\cite{chu2025sft}, where inference-time prefixes can deviate from the training distribution and cause errors to accumulate.
For reasoning, ReST~\cite{gulcehre2023reinforced} and STaR~\cite{zelikman2022star} self-train on sampled rationales verified by rewards or ground-truth answers.
In-context editing~\cite{qi2024context} shows that context-induced knowledge can be internalized through on-policy soft distillation.
Recent work includes SDPO~\cite{hubotter2026reinforcement}, which uses environment feedback as privileged information for learning, and OPSD~\cite{zhao2026self}, which uses ground-truth labels to provide reward signals for reasoning.
Uni-OPD~\cite{hou2026uni} further unifies OPD through a dual-perspective optimization recipe that jointly improves student exploration and teacher supervision reliability.
Nevertheless, most approaches still rely on a stronger teacher or verifiable ground-truth rewards.
Our work instead studies whether a single MLLM can supervise itself
using privileged visual evidence, without an external teacher,
ground-truth labels, or a verifier in the training objective.

\textbf{Fine-grained visual understanding for MLLMs.}
Large Language Models (LLMs) have demonstrated strong and robust capabilities in solving complex tasks~\citep{men2025shortgpt,yuan2025memsearcher,chen2025consistentchat,deepseekai2026deepseekv4,yang2025qwen3,vicuna2023}.
Built upon LLMs, Multimodal Large Language Models (MLLMs)~\cite{yuan2025shortv,mao2026pay,huang2026evidence,zhu2026ioupd,xiang2026teaching,openai2025gpt5} have demonstrated impressive capabilities in general visual understanding and reasoning.
Recent work on fine-grained multimodal understanding increasingly adopt a ``Thinking-with-Images'' strategy, where MLLMs gather visual evidence at inference time beyond a single forward pass~\cite{lai2025mini,wang2025pixel,zhang2025skywork,yu2025zoom,wei2025perception,zhang2025chain,zhou2024image,zhang2025finers,wei2025advancing,wu2025reinforcing,hou2025codev, wang2025vg,ma2026beyond,zhang2026instruction,zhang2025evaluating}.
DeepEyes series~\cite{zheng2025deepeyes,hong2025deepeyesv2} encourage visual tool calls such as ``Zoom in (Crop)'' and ``Search'' via reinforcement learning, while Thyme~\cite{zhang2025thyme} trains models to write code or manipulate visual inputs in pixel-space.
Training-free methods~\cite{liu2024chain,hu2024visual,shen2025zoomeye,liu2025hide,fu2025refocus,peng2025patch,luan2026textcot} take an alternative route by using tree search or attention-based localization to zoom into important regions during inference.
Although effective, these methods incur substantial inference cost and are less practical for real-time use.
Other approaches use specialized textual reasoning templates~\cite{wang2025vgr,wang2025traceable} or latent visual reasoning~\cite{yuan2025visual}, but require format-specific supervision and careful training.
ZwZ~\cite{wei2026zooming} uses RLVR (e.g., DAPO~\cite{yu2025dapo}) to improve single-pass fine-grained visual understanding without test-time tool use~\cite{mao2026thinking}.
Vision-OPD instead internalizes regional zooming into model parameters via self-distillation.

\section{Conclusion}

We introduce Vision-OPD, a simple and effective self-distillation framework for fine-grained visual understanding in MLLMs. 
The core idea is to let a model teach itself from privileged regional inputs. The teacher policy conditions on an evidence-centered crop, while the student policy observes the full image. 
By minimizing per-token divergence on the student's rollouts,
Vision-OPD provides dense supervision using a training objective that
does not require an external teacher, ground-truth labels, or a
verifier.
Experiments demonstrate that Vision-OPD substantially improves fine-grained understanding of existing MLLMs, surpassing much larger open-source models, closed-source models, and ``Thinking-with-Images'' agentic models.

\section{Limitations}

This work has several limitations. First, the training data are
constructed using automated object identification and segmentation,
model-generated questions, and filtering. This process may introduce
question-generation and data-selection biases, potentially limiting
the diversity of the resulting training samples.

Second, object identification and segmentation may produce inaccurate
region proposals or incomplete crops that omit relevant visual
evidence. Moreover, the crop-conditioned teacher may still produce
incorrect predictions even when the relevant evidence is visible,
leading to unreliable supervision. 

Finally, the experiments reported in this paper focus on Qwen3.5
models, and the generalization of Vision-OPD to non-Qwen model
families remains to be established.

\section*{Acknowledgment}
We sincerely thank the reviewers for their insightful comments and valuable suggestions. This work was supported by the National Key R\&D Program of China (2024YFC3308000), the Natural Science Foundation of China (No. 62476265, 62306303), Beijing Nova Program No. 202604841270.

\bibliographystyle{plainnat}
\bibliography{references}

\appendix

\section{Prompts}
\label{app:prompts}

\textbf{Prompts of benchmarks.}
To ensure reproducibility and facilitate future research, we provide here the complete set of prompts used to evaluate our models across all benchmarks.

\begin{tcolorbox}[
  title=V* Bench,
  fonttitle=\bfseries,
  colframe=black,
  colback=white,
  toptitle=1mm,
  bottomtitle=1mm,
  top=2mm,
  fontupper=\ttfamily\small,
  breakable,
]
\textless image\textgreater

\{question\}

\{options\}

Answer with the option's letter from the given choices directly.
\end{tcolorbox}

\begin{tcolorbox}[
  title=ZoomBench,
  fonttitle=\bfseries,
  colframe=black,
  colback=white,
  toptitle=1mm,
  bottomtitle=1mm,
  top=2mm,
  fontupper=\ttfamily\small,
  breakable,
]
\textless image\textgreater

\{question\}

\{options\}
\end{tcolorbox}

\begin{tcolorbox}[
  title=HR-Bench,
  fonttitle=\bfseries,
  colframe=black,
  colback=white,
  toptitle=1mm,
  bottomtitle=1mm,
  top=2mm,
  fontupper=\ttfamily\small,
  breakable,
]
\textless image\textgreater

\{question\} Select from the following choices.

\{options\}
\end{tcolorbox}

\begin{tcolorbox}[
  title=MME-RealWorld EN,
  fonttitle=\bfseries,
  colframe=black,
  colback=white,
  toptitle=1mm,
  bottomtitle=1mm,
  top=2mm,
  fontupper=\ttfamily\small,
  breakable,
]
\textless image\textgreater

\{question\} The choices are listed below:

\{options\}

Select the best answer to the above multiple-choice question based on the image. Respond with only the letter (A, B, C, D, or E) of the correct option.

The best answer is:
\end{tcolorbox}

\begin{tcolorbox}[
  title=MME-RealWorld CN,
  fonttitle=\bfseries,
  colframe=black,
  colback=white,
  toptitle=1mm,
  bottomtitle=1mm,
  top=2mm,
  fontupper=\small,
  breakable,
]
\begin{CJK*}{UTF8}{gbsn}
\textless image\textgreater

\{question\} 选项如下所示:

\{options\}

根据图像选择上述多项选择题的最佳答案。只需回答正确选项的字母（A, B, C, D 或 E）。

最佳答案为：
\end{CJK*}
\end{tcolorbox}

\begin{tcolorbox}[
  title=MMVP,
  fonttitle=\bfseries,
  colframe=black,
  colback=white,
  toptitle=1mm,
  bottomtitle=1mm,
  top=2mm,
  fontupper=\ttfamily\small,
  breakable,
]
\textless image\textgreater

\{question\} \{options\}
\end{tcolorbox}

\begin{tcolorbox}[
  title=CV-Bench,
  fonttitle=\bfseries,
  colframe=black,
  colback=white,
  toptitle=1mm,
  bottomtitle=1mm,
  top=2mm,
  fontupper=\ttfamily\small,
  breakable,
]
\textless image\textgreater

\{question\} Select from the following choices.

\{options\}
\end{tcolorbox}

\begin{tcolorbox}[
  title=MMStar,
  fonttitle=\bfseries,
  colframe=black,
  colback=white,
  toptitle=1mm,
  bottomtitle=1mm,
  top=2mm,
  fontupper=\ttfamily\small,
  breakable,
]
\textless image\textgreater

\{question\}

Options: \{options\}
\end{tcolorbox}

\begin{tcolorbox}[
  title=POPE,
  fonttitle=\bfseries,
  colframe=black,
  colback=white,
  toptitle=1mm,
  bottomtitle=1mm,
  top=2mm,
  fontupper=\ttfamily\small,
  breakable,
]
\textless image\textgreater

\{question\}

Answer the question using a single word or phrase.
\end{tcolorbox}

\textbf{Prompt for question generation.}
For each retained region, we use Qwen3.5-397B to generate
fine-grained questions based on the cropped image. 

\begin{tcolorbox}[
  title=Question Generation Prompt,
  fonttitle=\bfseries,
  colframe=black,
  colback=white,
  toptitle=1mm,
  bottomtitle=1mm,
  top=2mm,
  fontupper=\ttfamily\small,
  breakable,
]
\textless cropped image\textgreater

Reference questions:

\{randomly sampled seed questions\}

Using only the provided image, generate three diverse visual
questions. Each question should be answerable from the image and
have a concise, factual answer.

Questions may cover object identity, text, counting, spatial
relationships, shape, material, or other visible details.

If the image does not support meaningful questions, return an empty
JSON list: [].

Otherwise, return a JSON list containing three objects, each with
a ``question'' field:

[
  \{ ``question'': ``\textless question 1\textgreater'' \},
  \{ ``question'': ``\textless question 2\textgreater'' \},
  \{ ``question'': ``\textless question 3\textgreater'' \}
]
\end{tcolorbox}

\section{Implementation details of data synthesis}
\label{app:data_synthesis}

The pipeline
constructs paired visual contexts for each fine-grained question:
a global image $x$ for the student and an evidence-centered crop
$x'$ for the teacher.

We use images from SA-1B~\citep{kirillov2023segment} to synthesize
the training dataset. Given a raw image $I$, we first use
Qwen3.5-397B to identify objects in the image. For each identified
object, SAM3~\citep{carion2025sam} generates a segmentation mask and
a corresponding bounding box $B$. 
We only retain regions whose SAM3
confidence score is greater than $0.75$.

For each retained bounding box, we first crop the corresponding
region from the original image $I$ and resize it by $2\times$ to
obtain a regional image $R$. Qwen3.5-397B receives $R$, without
the full image, and generates fine-grained questions intended to be
answerable from the crop alone. 

We ground each crop-level question back to the full image by
overlaying its bounding box $B$ on the original image $I$. The
resulting image is the student input $x$. We also add a spatial
constraint to the question, such as ``Only focus on the objects
inside the red bounding box,'' to indicate the relevant region.

To construct the teacher input $x'$, we crop the corresponding region
from the red-box image $x$ and resize it by $2\times$. This is a
separate step from the crop of the original image used for question
generation. The resulting paired inputs present the question under
two visual conditions: the student observes the full image with
spatial guidance, whereas the teacher observes the enlarged regional
crop. The red box and spatial constraint help reduce referential
ambiguity when the crop-level question is presented with the full
image.

To support comparisons with label-dependent training strategies,
including off-policy distillation, RLVR, and
OPSD~\citep{zhao2026self}, we use Qwen3.5-397B as an answer
generator. For each crop--question pair, we sample multiple
responses conditioned on the original-image crop $R$. We retain a
sample only when its majority answer reaches a consensus score
greater than $0.75$. The retained questions are formatted as
multiple-choice prompts with candidate options. In total, the
resulting training dataset contains approximately 6.2K samples.

The generated answer labels are used for the label-dependent
baselines. Consensus filtering also determines which examples enter
the shared training dataset; however, the answer labels are not
used as targets in the Vision-OPD distillation objective.

\section{Case Study}
\label{app:case_study}
\label{sec:failure-cases}

\textbf{Fine-grained visual understanding capability of Vision-OPD.}
As shown in Table~\ref{tab:case_boat_number}, Vision-OPD-9B answers
correctly while Qwen-3.5-9B fails, demonstrating that Vision-OPD can
effectively internalize fine-grained visual understanding capabilities
into existing MLLMs.

\textbf{Crop--question mismatch and unreliable supervision.}
Table~\ref{tab:failure-water} shows a failure case in which the
crop does not contain the relevant part of the image. The question asks about
the water at the bottom edge of the full image, but the crop primarily
contains the beach and only a narrow strip of water. Given the full image,
Qwen3.5-9B answers ``C. blue,'' which is consistent with the visible
blue-green water; given the teacher input, the same model answers ``D. brown.''
This example illustrates how a mismatched crop can produce an
unreliable crop-conditioned response and potentially weaken the supervision
available to the student.

\begin{table}[t]
  \begin{minipage}{0.99\linewidth}
\centering
\captionof{table}{Vision-OPD reads a small number written on a distant boat.}
\resizebox{\textwidth}{!}
{%
\begin{tabular}{p{2cm} p{16.75cm}}
\toprule
 \multicolumn{2}{l}{\bf Visual input example, Small Text/Number Reading:}  \\
\midrule
&  \includegraphics[height=18.25cm]{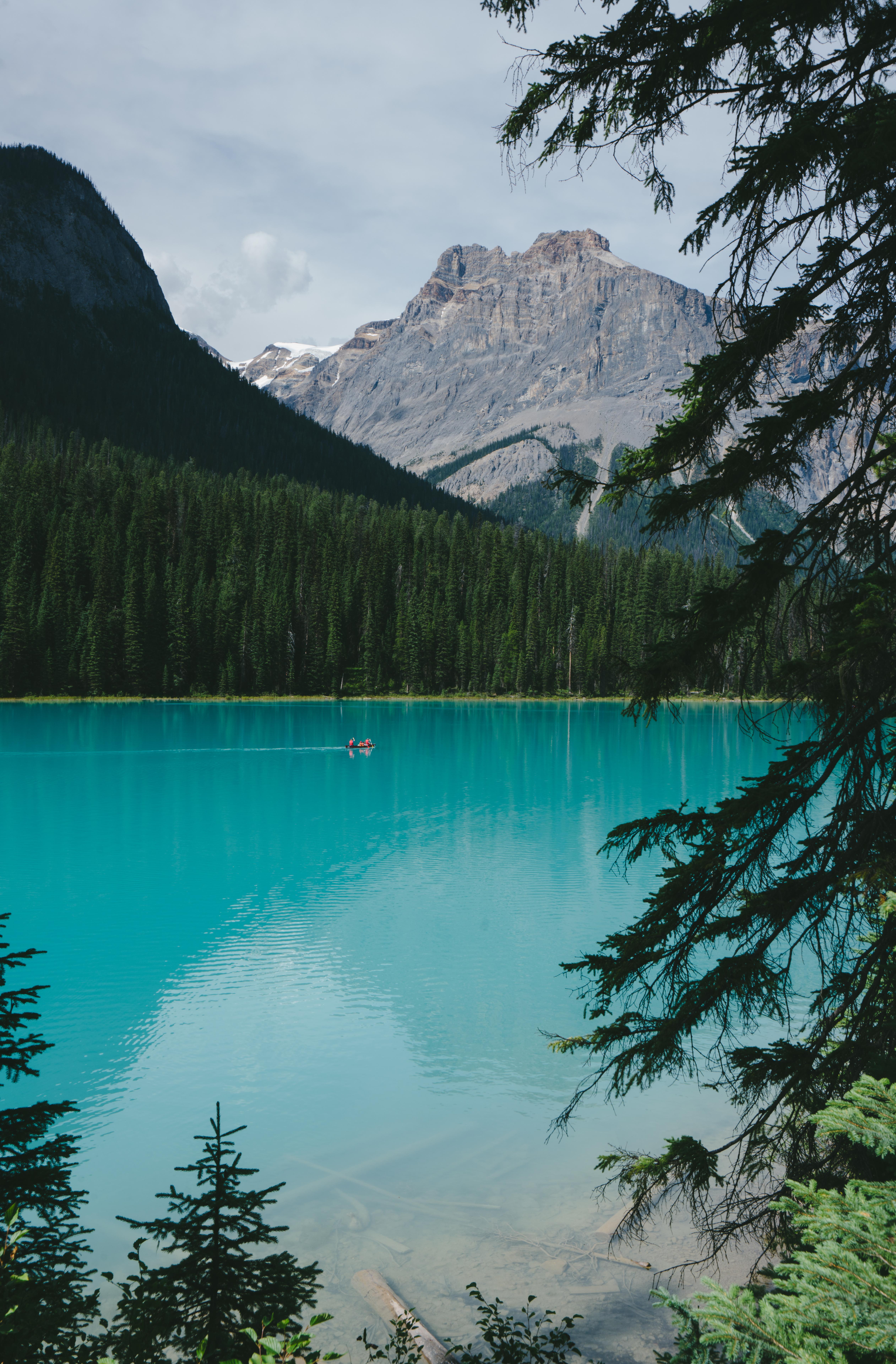} \\

User & What's the number written on the boat? Select from the following choices.\newline (A) 10\newline (B) 25\newline (C) 15\newline (D) 5 \\
\midrule
Qwen-3.5 & The image shows a small boat with people on a turquoise lake, surrounded by mountains and trees. However, **there is no visible number written on the boat** in the image.\par Looking closely at the boat (which appears to be a canoe or small rowboat), it's too distant and low-resolution to discern any markings or numbers. The question asks us to select from given choices: \newline (A) 10  \newline (B) 25  \newline (C) 15  \newline (D) 5  \par Since none of these numbers are visibly present on the boat in the image, and assuming this is a trick or observation-based question, the most reasonable conclusion is that **no number is visible**, but since we must choose from the options, perhaps the intended answer is based on an external context or misinterpretation.\par However, if this is a standard multiple-choice question from a known source (like a quiz or exam), and given that **canoes in some parks or regions are numbered for identification**, and **number 5** is commonly used in such contexts -- especially if this is Lake Louise or similar -- then **(D) 5** might be the expected answer.\par But strictly speaking, **based solely on the visual content of the image provided**, there is **no number visible on the boat**.\par ---\par Given the constraints and typical test design, the best guess is:\par {\color{red}**(D) 5**}\par *(Note: This is an inference based on common practices, not direct visual evidence.)* \\
\midrule
Vision-OPD & Looking at the image, there is a small boat on the turquoise lake with people in it. On the side of the boat, near the front (bow), there appears to be a number painted or stenciled.\par The number visible is **``15''** -- it's white and clearly legible against the darker hull.\par Therefore, the correct answer is:\par **(C) 15**\par Final Answer: {\color{green}**C**} \\
\bottomrule
\end{tabular}
}
\label{tab:case_boat_number}
  \end{minipage}
\end{table}

\begin{table}[t]
    \centering
    \small
    \caption{An example illustrating a failed proposal,
    question--crop mismatch and an unreliable
    crop-conditioned response.}
    \label{tab:failure-water}
    \setlength{\tabcolsep}{4pt}
    \begin{tabular}{p{0.19\linewidth} p{0.74\linewidth}}
        \toprule
        \multicolumn{2}{l}{\textbf{Full-image input with the retained box:}} \\
        & \includegraphics[width=\linewidth]{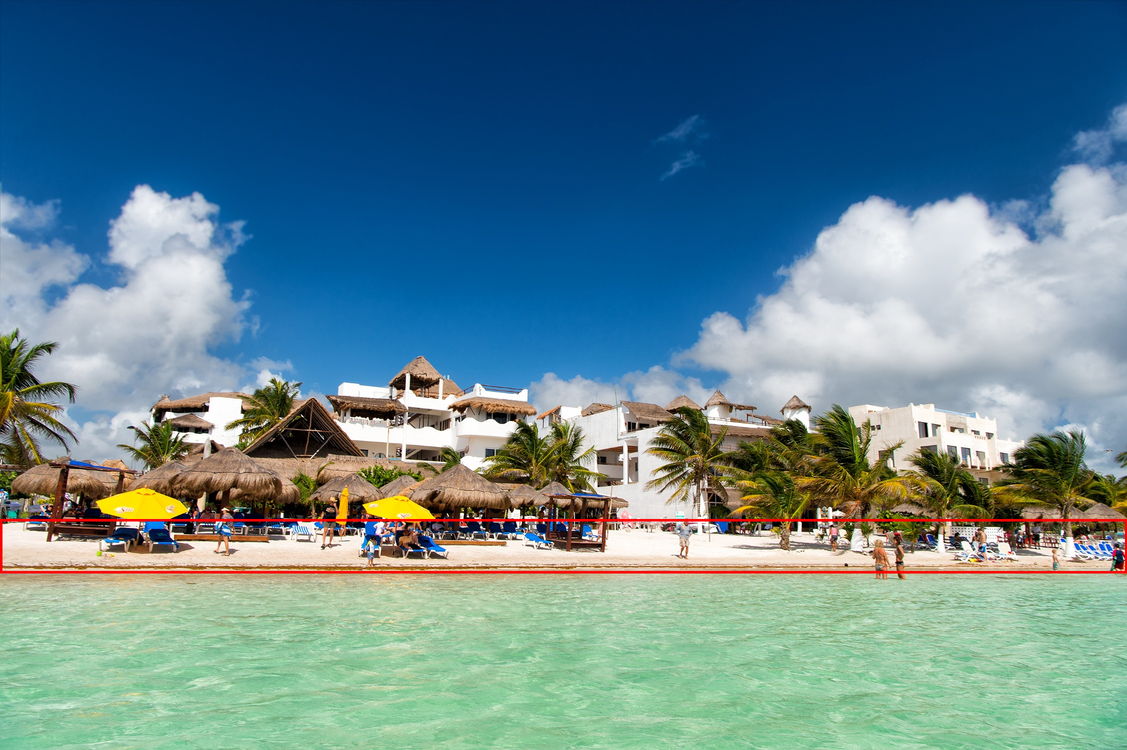} \\

        \multicolumn{2}{l}{\textbf{Corresponding teacher input:}} \\
        & \includegraphics[width=\linewidth]{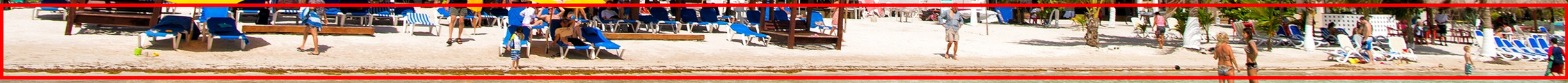} \\

        \midrule
        Question & What is the color of the water at the bottom edge of the image?
        \newline Only focus on the objects inside the red bounding box in the
        image to answer this question.
        \newline A. green \quad B. clear \quad C. blue \quad D. brown
        \newline Answer with the option's letter from the given choices. \\

        \midrule
        Qwen3.5 (student) & C. blue. \\
        Qwen3.5 (teacher) & D. brown. \\
        \bottomrule
    \end{tabular}
\end{table}

\FloatBarrier

\end{document}